\documentclass{article}
\usepackage{arxiv}
\usepackage[utf8]{inputenc} % allow utf-8 input
\usepackage[T1]{fontenc}    % use 8-bit T1 fonts
\usepackage{hyperref}       % hyperlinks
\usepackage{url}            % simple URL typesetting
\usepackage{booktabs}       % professional-quality tables
\usepackage{amsfonts}       % blackboard math symbols
\usepackage{nicefrac}       % compact symbols for 1/2, etc.
\usepackage{microtype}      % microtypography
\usepackage{lipsum}

\usepackage{cite}
\usepackage{amsmath,amssymb}
\usepackage{algorithmic}
\usepackage{graphicx}
\usepackage{textcomp}
\usepackage{xcolor}
\usepackage{multirow}
\usepackage{subfigure}

\title{A Tiny Machine Learning Model for Point Cloud Object Classification}

\author{Min Zhang \thanks{The first two authors contributed equally to this work. Corresponding author: Min Zhang, zhan980@alumni.usc.edu.} \\ Media Communications Lab\\ University of Southern California\\ Los Angeles, CA, USA\\ \And Jintang Xue $^*$\\ Media Communications Lab\\ University of Southern California\\ Los Angeles, CA, USA\\ \And Pranav Kadam\\ Media Communications Lab\\ University of Southern California\\ Los Angeles, CA, USA\\ \And Hardik Prajapati\\ Media Communications Lab\\ University of Southern California\\ Los Angeles, CA, USA\\ \And Shan Liu \\ Tencent Media Lab\\ Tencent America\\ Palo Alto, CA, USA\\ \And C.-C. Jay Kuo\\ Media Communications Lab\\ University of Southern California\\ Los Angeles, CA, USA}

\begin{document}
\maketitle
\begin{abstract}
The design of a tiny machine learning model, which can be deployed in mobile and edge devices, for point cloud object classification is investigated in this work. To achieve this objective, we replace the multi-scale representation of a point cloud object with a single-scale representation for complexity reduction, and exploit rich 3D geometric information of a point cloud object for performance improvement. The proposed solution is named Green-PointHop due to its low computational complexity. We evaluate the performance of Green-PointHop on ModelNet40 and ScanObjectNN two datasets.  Green-PointHop has a model size of 64K parameters. It demands 2.3M floating-point operations (FLOPs) to classify a ModelNet40 object of 1024 down-sampled points. Its classification performance gaps against the state-of-the-art DGCNN method are 3\% and 7\% for ModelNet40 and ScanObjectNN, respectively. On the other hand, the model size and inference complexity of DGCNN are 42X and 1203X of those of Green-PointHop, respectively. 
\end{abstract}

\keywords{Point clouds \and object classification \and tiny learning model \and green learning \and PointHop}

\section{Introduction}\label{sec:introduction}

Given a point cloud object scan, the goal of point cloud classification is to assign it a category label. Point cloud object classification is a fundamental task in point cloud analysis and processing. It lays the foundation for other advanced 3D vision tasks such as semantic/instance segmentation, object detection, registration and generation. Different from semantic segmentation or object detection, which needs to deal with large-scale and noisy point clouds, point cloud object classification often targets at classifying small-scale objects that are clean and well aligned. The point cloud object classification task relies on the understanding of the local and global structures of an object. Moreover, efficiency of a classifier matters in practical 3D real-time systems that often contain a few advanced tasks and applications.  The design of a tiny machine learning model to be deployed in mobile and edge devices is our focus in this work. 

Two families of machine learning models have been proposed for 3D point cloud object classification. The first one relies on the end-to-end optimized deep learning (DL) technique. A representative method is PointNet \cite{qi2017pointnet}. The second one adopts the green learning (GL) methodology \cite{kuo2022green} aiming at smaller model sizes and lower computational complexity. An illustrative example is PointHop \cite{zhang2020pointhop}. To design a tiny point cloud classifier, we have observations on methods of these two families below. 

As a pioneering work, PointNet has influenced DL-based neural networks on point cloud classification significantly. Its follow-ups include \cite{hu2020randla, jiang2018pointsift, li2018pointcnn, qi2017pointnet++, wang2019dynamic}. All of them attempt to learn a richer context.  Recently, more complex models \cite{guo2021pct, zhao2021point} are proposed to push the performance furthermore. Their computational complexity and memory costs increase accordingly. On the other hand, we observe that the performance of point cloud classifiers is not much affected by their layer depth and complexity. For instance, PointNet achieves 47.6\% mIoU (the mean intersection over union) in semantic segmentation for the S3DIS dataset \cite{armeni_cvpr16} while Point Transformer \cite{zhao2021point} increases the mIoU value to 73.5\%. In contrast, PointNet and Point Transformer achieve 89.2\% and 93.7\% classification accuracy on the ModelNet40 dataset \cite{wu20153d}, respectively. As compared with the 25.9\% performance gain in semantic segmentation, the improvement of 4.5\% classification accuracy of Point Transformer over PointNet is not impressive. 
 
Solutions to point cloud classification, segmentation and registration can be designed using the GL methodology to achieve lower inference complexity and smaller model sizes.  Since no backpropagation is used to determine model parameters, their training is also very fast. All of these advantages are enabled by the unsupervised representation learning module in GL. PointHop is the first GL-based method on point cloud classification. It learns local-to-global attributes hierarchically. The framework has been successfully extended to other tasks such as point cloud segmentation \cite{zhang2020unsupervised} and point cloud registration \cite{kadam2020unsupervised}. Nevertheless, the limited performance gain in classification accuracy (i.e., 2\%) from PointHop to its enhanced solution, PointHop++ \cite{zhang2020pointhop++}, is also observed.

Based on the above observations, there appears to be little gain in building a complex point cloud classifier.  Although GL-based point cloud classifiers are more efficient than their DL-based counterparts, we wonder whether it is possible to obtain an even more economical model. Models of both DL-based and GL-based families share one common principle -- all of them represent point cloud objects in multiple scales. To reduce the complexity of existing solutions, we replace the multi-scale representation of a point cloud object with a single-scale representation for complexity reduction, and exploit rich 3D geometric information of a point cloud object for performance preservation. The proposed solution is named Green-PointHop due to its low computational complexity. 

Since Green-PointHop has only one hop, its computation time and model size can be reduced. It compensates the performance loss of a shallow model by concatenating 3D complementary geometric information using the global, cone and inverted cone aggregations. As compared to PointHop and PointHop++, which need four hops to learn hierarchical representations of point clouds, Green-PointHop has a much simpler architecture. We evaluate the performance of Green-PointHop on ModelNet40 and ScanObjectNN two datasets. Green-PointHop has a model size of 64K parameters. It demands 2.3M floating-point operations (FLOPs) to classify a ModelNet40 object of 1024 down-sampled points. Its classification performance gaps against the state-of-the-art DGCNN method are 3\% and 7\% for ModelNet40 and ScanObjectNN, respectively. On the other hand, the model size and inference complexity of DGCNN are 42X and 1203X of those of Green-PointHop, respectively. 

The rest of the paper is organized as follows. Related work is reviewed in Sec. \ref{sec:review}. The Green-PointHop method is proposed in Sec. \ref{sec:method}. Experimental results are shown in Sec. \ref{sec:experiments}. Finally, concluding remarks are given in Sec. \ref{sec:conclusion}. 

\section{Related Work} \label{sec:review}

\subsection{Methods for Point Cloud Analysis}

For traditional point cloud analysis methods, it is common to use local geometric properties as point cloud representations. Representative 3D local geometric descriptors include PFH \cite{rusu2009fast}, FPFH \cite{rusu2009fast}, ISS \cite{zhong2009intrinsic}, and SHOT  \cite{tombari2010unique}, etc. Traditional point cloud analysis methods are mathematically interpretable. They offer good and fast results in simple scenarios. Yet, they are not effective in handling complex objects. 

With the success of DL in image processing and computer vision, researchers tried to apply DL to point clouds but encountered difficulty due to the unordered and unstructured nature of 3D points initially. As a pioneering DL solution, PointNet solves this problem using multi-layer perceptrons (MLPs) and max aggregation. Since the max aggregation function is a symmetric one, PointNet is invariant to the order of points. PointNet only learns the global structure of point clouds without considering local structures of points. 

Its follow-ups, e.g., PointNet++ \cite{qi2017pointnet++}, PointCNN \cite{li2018pointcnn}, PointSIFT \cite{jiang2018pointsift}, DGCNN \cite{wang2019dynamic}, RandLANet \cite{hu2020randla}, learn richer local contexts. PointNet++ applies PointNet to the local neighborhood of each point and aggregates local features hierarchically. PointCNN uses a $\chi$-Conv to aggregate features in a local region with a latent and potentially canonical order. Inspired by the SIFT descriptor \cite{lowe2004distinctive}, PointSIFT uses an orientation-encoding unit to encode eight orientations. DGCNN \cite{wang2019dynamic} updates local regions at each layer using point features (instead of 3D coordinates). This idea works better in capturing the semantic information. RandLANet \cite{hu2020randla} has an attention mechanism and aggregates features with attention scores. 

DL-based methods achieve good performance in complex tasks but require a large number of training data, expensive annotation, and huge computational resources. Moreover, they are not as transparent as traditional methods.  GL-based point cloud analysis methods strike a balance between simple traditional methods and costly DL-based methods.

PointHop \cite{zhang2020pointhop} and PointHop++ \cite{zhang2020pointhop++} are two pioneering GL-based point cloud processing systems. PointHop++ is an improved version of PointHop method. They exploit successive subspace learning (SSL) \cite{chen2020pixelhop} to learn rich representations of point clouds. Specifically, they adopt an unsupervised feature extraction module without any class labels. They have an extremely low training complexity since no backpropagation is used. They only demand one feedforward pass to learn model parameters in the feature extraction module. 

PointHop and PointHop++ lay the foundation for other advanced GL-based
point cloud methods. The representation learning technique proposed in
PointHop is commonly adopted by others.  Examples include point cloud
classification \cite{zhang2020pointhop, zhang2020pointhop++,
kadam2023s3i}, small object part segmentation
\cite{zhang2020unsupervised}, large-scale semantic segmentation
\cite{zhang2021gsip}, registration \cite{kadam2020unsupervised,
kadam2021r}, odometry \cite{kadam2021gpco}, joint object retrieval and
pose estimation \cite{kadam2022pcrp}, rotational-invariant object
classification \cite{kadam2023s3i}, and scene flow estimation
\cite{kadam2023pointflowhop}. 

\subsection{GL Methodology and Applications}

GL was originally developed for image classification tasks \cite{chen2020pixelhop, chen2020pixelhop++, yang2021pixelhop} based on a sequence of efforts in analyzing the operations of convolutional neural networks (CNNs) \cite{chen2018saak, kuo2016understanding, kuo2018data, kuo2022green, kuo2019interpretable}. The GL paradigm has been successfully applied to other 2D vision tasks such as face biometrics \cite{rouhsedaghat2020facehop}, deepfake detection \cite{chen2021defakehop}, anomaly detection \cite{zhang2021anomalyhop}, image generation \cite{lei2022genhop, lei2020nites, lei2021tghop}, and object tracking \cite{zhou2021uhp, zhou2022gusot, zhou2022uhp, zhou2021unsupervised}. The GL paradigm has also been exteneded to 3D vision problems, where a wide range of point cloud processing algorithms have been developed in recent years \cite{liu20213d}. 

\section{Green-PointHop Method} \label{sec:method}

\subsection{Motivation and System Overview} \label{sec:motivation}

It is common to use an hierarchical processing pipeline for object recognition in 2D images, say, the multi-layer network architecture in the context of CNNs. The computational neurons at shallower (or deeper) layers have smaller (or larger) receptive fields. More neurons are needed at deeper layers due to higher content diversity in a larger receptive field.  Although the same principle holds for 3D point clouds, there is a major difference in pooling operations between images and point cloud data. 

A (2x2)-to-(1x1) local pooling is applied to filter responses in hierarchical image processing.  This is proper due to the structured order of image pixels. All relative 2D spatial information is still preserved in pooled responses. Yet, the situation changes in 3D point clouds. To deal with unstructured 3D points, we need to define two operations, pooling and aggregation, separately. The pooling operation is applied to input 3D points (rather than filter responses) while the aggregation function, which has to be a symmetric one, is applied to filter responses. The underlying 3D spatial information is lost after aggregation. The hierarchical architecture computes spectral representations at different scales. Its cost becomes higher as the layer goes deeper. 

To reduce the computational and memory costs, we consider a single-hop model called Green-PointHop, where no pooling is used to reduce the point number. It computes the spectral representation of a local neighborhood centered at each point using the Saab transform \cite{kuo2019interpretable}, and conducts aggregations across points of an object and its partial views (or sub-objects). This new yet simple idea has a clear geometric interpretation. Furthermore, its model size and computation complexity can be significantly reduced since it has filtering operations at a single scale. 

\begin{figure}[htbp]
\centering
\centerline{\includegraphics[width=\linewidth]{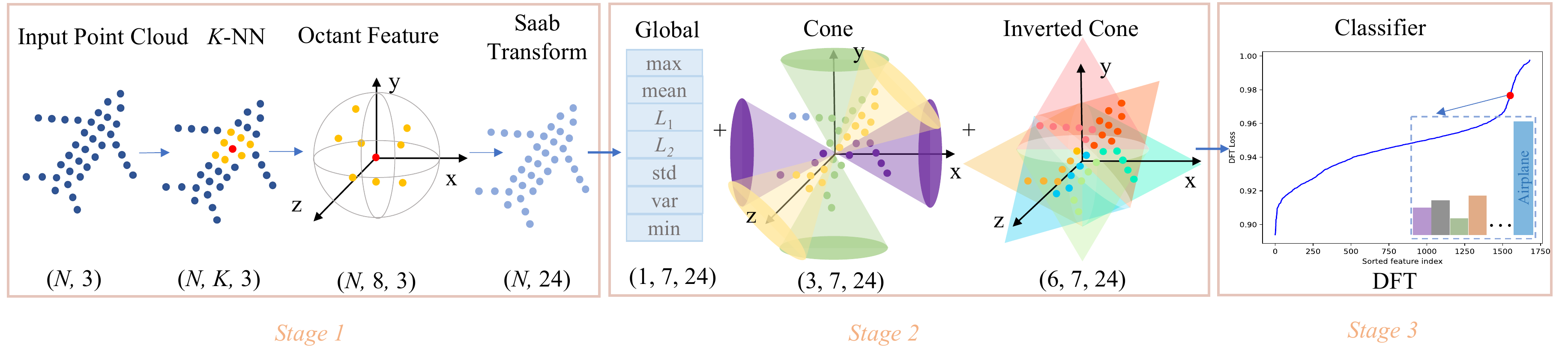}}
\caption{An overview of the proposed Green-PointHop method. It consists of three stages. In the first stage, it computes a point-wise representation for every point in the input point cloud. For example, the neighborhood of the red point is marked in yellow and determined by the KNN search. Then, the point-wise representation vector is derived using the eight-octant partitioning and averaging, which are followed by Saab filtering. In the second stage, seven schemes (i.e., max, mean, $L_1$, $L_2$, std, var, and min) are used to aggregate point-wise representation vectors into global and regional representation vectors. Regions are defined by points inside cones or inverted cones. In the third stage, discriminant features are selected using the discriminant feature test (DFT) and object classification is performed using the linear-least-squares-regression (LLSR) classifier. Unsupervised learning is adopted in the first and the second stages, where no object labels are used. Supervised learning is used in the third stage.} 
\label{fig:pipeline}
\end{figure}

An overview of the proposed Green-PointHop system is shown in Fig. \ref{fig:pipeline}. The input point cloud scan consists of $N$ points. Each point, $p_i$, has three Cartesian coordinates $(x_i, y_i, z_i)$, representing its location in the 3D space. To obtain a discriminant point cloud representation for classification, we propose a three-stage design. In the first stage, we compute local representations for each point as detailed in Sec. \ref{sec:3.1}. In the second stage, we perform geometric aggregations of point-wise representations in Sec. \ref{sec:3.2}. In the third stage, we rank the discriminant power of each representation dimension from the highest to the lowest, select discriminant ones based on the elbow point of the curve, and feed them into a classifier to obtain an object label. The detail is given in Sec. \ref{sec:3.3}. 

%%%%%%%%%%%%%%%%%%%%%%%%%%%%%%%%%%%%%%%%%%%%%%%%%%%%%%%
%%%%%%%%%%%%%%%%%%%%%%%%%%%%%%%%%%%%%%%%%%%%%%%%%%%%%%%%

\subsection{Stage 1: Point-wise Representation Learning}\label{sec:3.1}

{\em Raw Point-wise Representation Vector.} For each point, we find its $K$ nearest neighbors using the $K$-NN search. As shown in the first stage of Fig. \ref{fig:pipeline}, the red point is the target point while the yellow points are its $K$ nearest neighbors. We compute local coordinates of yellow points with the red point as the origin, partition them using eight octants, and calculate the centroid of points inside each octant. If an octant does not have any point, its centroid is set to $(0,0,0)^T$. Since there are 8 octants and each centroid has 3 spatial coordinates, we have a raw representation vector of 24 dimensions. 

{\em Saab Filtering.} There are correlations between dimensions of raw point-wise representation vector. It is advantageous to decorrelate them using the Saab transform based on our prior experiences in PointHop and PointHop++. The Saab transform \cite{kuo2019interpretable} contains one DC kernel (or filter) and 23 AC kernels (or filters). The DC kernel is a constant-element vector.  Since DC-removed vectors can be treated as zero-mean vectors, we conduct the principal component analysis (PCA) on them and obtain 23 AC kernels.  Then, by multiplying the Saab transform matrix of dimension $24 \times 24$ with the raw representation vector of dimension 24, we obtain 24 filter responses, which define the ultimate point-wise representation vector. 

The point-wise representation vector was first proposed in \cite{zhang2020pointhop}. It has three nice properties. First, it is invariant under index permutation of points. Second, it is a spectral representation via the Saab transform, which is more effective than the raw representation in the spatial domain. Third, it is a representation derived by unsupervised learning. No object labels are needed. 

\subsection{Stage 2: Global and Regional Representation Learning} \label{sec:3.2}

Point-wise representation vectors alone are too local to be discriminant. Besides, the number of these vectors is proportional to the number of points, which is too many to feed to the classifier directly. It is essential to aggregrate point-wise representation vectors across a set of points to offer robust and discriminant representations of manageable sizes. The aggregation of point-wise octant representation vectors is proven to be effective in prior GL-based point cloud analysis work such as PointHop, PointHop++ and R-PointHop \cite{kadam2021r}. Some adjustments are made to tailor to Green-PointHop as described below. 

{\em Aggregation Functions.} The aggregation function is applied to individual elements of the representation vectors of selected points. It should be a symmetric function so that the aggregation output is invariant to the permutation of point indices. The maximum (or max) aggregation function was used in PointNet. Three more aggregation functions are added in PointHop and PointHop++; namely, the mean, the $L_1$-norm, and the $L_2$-norm. They aggregate point-wise representation vectors in a complementary manner and offer better performance. Here, we add three more aggregation functions to Green-PointHop: the minimum (min), the standard deviation (std) and the variance (var). Thus, Green-PointHop has seven aggregation functions in total as shown in the second stage of Fig.  \ref{fig:pipeline}. The effectiveness of these aggregation functions is analyzed in Sec. \ref{ablation_study}. 

{\em Aggregation Regions.} The aggregation function has to work on a set of points. One default choice is to include all points of an input point cloud. It is called the global aggregation. Furthermore, we can consider a subset of points. We use the centroid of all points as the new origin while keeping the same pointing directions of the x, y, and z three axes in the following discussion. We define a symmetric cone-shaped region by including all points whose angles with respect to the positive or negative x-axis are less than a predefined parameter, $\theta_1$. This cone region is shown in purple in the second stage of Fig. \ref{fig:pipeline}. Similarly, we define two more symmetric cone-shaped regions with respect to the y- and z-axes.  They are colored in green and yellow, respectively. We can further manipulate the cone that covers points in the positive x-axis as follows. We move the vertex of the cone to $(\Delta, 0, 0)^T$, where $\Delta > 0$ is a parameter selected by users and invert the cone orientation by 180 degrees. Then, points with the new vertex as the origin with angles against the negative x-axis less than $\theta_2$ define an inverted cone region that has its base on the y-z plane (i.e., $x=0$). With the same manipulations, we can obtain 6 inverted cone-shape regions. They are marked by 6 different colors in the right of the second module in Fig. \ref{fig:pipeline}.  In the experiments, we set $\theta_1 = 75^{\circ}$ for three symmetric cone-shaped regions and $\theta_2 = 45^{\circ}$ for six inverted cone-shaped regions.  The former three are used to capture symmetrical shape properties while the latter six are used to capture non-symmetric shape characteristics. 

To summarize, there are 10 regions in our design (i.e., one global, three symmetric cones and six inverted cones). We apply the 7 aggregation functions to the 24D point-wise representation vector of all points in each region, and concatenate all of them to form a joint representation vector of dimension $7 \times 24 \times 10=1680$, which serves as the input to the last stage. 

%%%%%%%%%%%%%%%%%%%%%%%%%%%%%%%%%%%%%%%%%%%%%%%%%%%%%%%
\begin{figure}[htbp!]
\centerline{\includegraphics[width=2.2in]{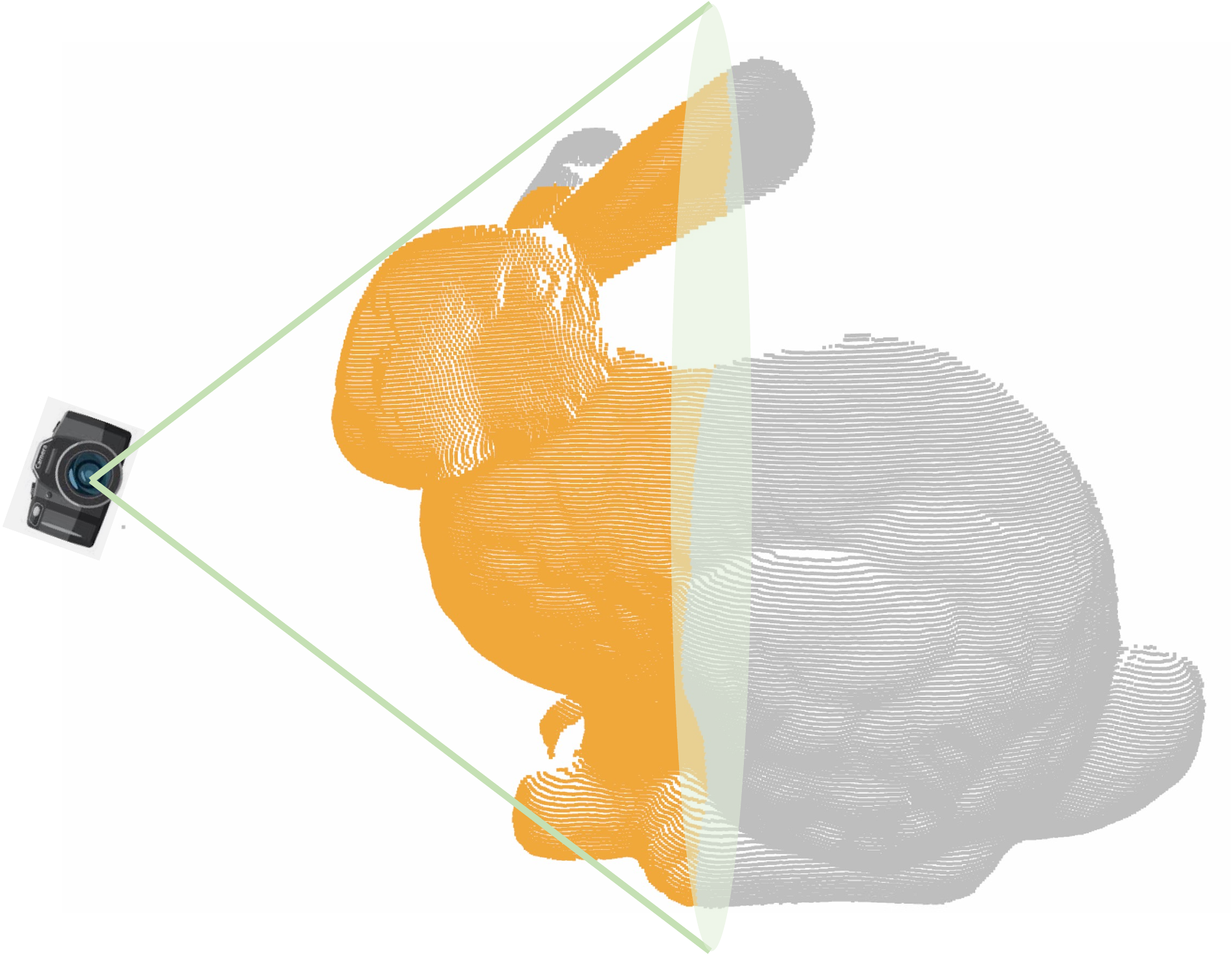}}
\caption{An illustration for the intuition behind the regional cone-shaped aggregation. A pinhole camera is set at a certain distance to a Stanford bunny and views a specific part of it.}\label{fig:bunny}
\end{figure}
%%%%%%%%%%%%%%%%%%%%%%%%%%%%%%%%%%%%%%%%%%%%%%%%%%%%%%%%

There is an intuitive way to explain the regional cone-shaped design. Each cone can be treated as a pinhole camera at different locations viewing a specific part of the object. For example, the pinhole camera is at the center of the object and views parts of the object along each axis. The viewing angle has a certain limitation bounded by $\theta_1$. As an alternative, the camera goes to different sides of the object, turns around, and examines the object from each side to the center, where the viewing angle is bounded by $\theta_2$. In Fig. \ref{fig:bunny}, the Stanford bunny \cite{turk1994zippered} is used to illustrate our intuition. 

\subsection{Stage 3: Feature Selection and Decision Learing}\label{sec:3.3}

{\em From Representation Vectors to Features.} As stated above, the total dimension number from Stage 2 is 1680. We can select a subset of them to serve as the final feature set. Here, we adopt a supervised feature selection method called the discriminant feature test (DFT) \cite{yang2022supervised}. The DFT is a tool used to measure the discriminant power of each dimension independently. It consists of two steps: 1) The weighted entropy loss for each 1D dimension is calculated at a partition, which is chosen from a set of uniformly partitioned points. 2) The lowest weighted entropy loss is set to the DFT loss of the dimension. Then, we sort dimensions from the lowest to the highest DFT loss values as shown in the third stage of Fig. \ref{fig:pipeline}, where the x-axis is the sorted dimension index and the y-axis is the DFT loss value. The lower the DFT loss, the higher the discriminant power. Then, we can use the elbow point of the curve as indicated by the red point to determine a subset of discriminant features. Features with their loss values less than that of the red point are selected. 

{\em Classifier Selection.} We try a couple of classifiers and find that the linear-least-squares-regression (LLSR) classifier gives the best tradeoff between the classification performance, computational complexity, and the number of model parameters. Thus, we choose it as the classifier for Green-PointHop. 

\section{Experiments} \label{sec:experiments}

\subsection{Experimental Setup}

We first evaluate Green-PointHop on ModelNet40 \cite{wu20153d}. It has 40 object categories. The dataset contains 9,843 training samples and 2,468 test samples. We randomly down-sample each point cloud scan from 2,048 points to 1,024 points for further processing. In experiments, we set $K=32$ in the KNN search, select 1569 features using DFT, and feed them into the LLSR classifier. Exhaustive experimental results such as classification accuracy, model sizes, and time complexity in inference, and ablation study are reported in Sec. \ref{subsec:ModelNet40}. 

To show the generalizability of Green-PointHop, we also evaluate its performance on the ScanObjectNN dataset \cite{uy2019revisiting}. It consists of objects extracted from real world scenes. The dataset has several scenarios such as the "object only", "object with background", and "object with perturbation". To align with ModelNet40, we evaluate the "object only" scenario only. For this scenario, we have 2,309 training samples and 581 test samples from 15 object classes. Since real world models have background noise and incomplete scans, we adjust some hyper-parameters in Green-PointHop. In experiments, we set $K=48$ in the KNN search, $65^{\circ}$ for all cones, and select 1108 features based on DFT.  The experimental results are given in Sec. \ref{ScanObjectNN}. 

\subsection{ModelNet40 Dataset}\label{subsec:ModelNet40}

\subsubsection{Classification Performance}

We compare the classification accuracy results of several methods on ModelNet40 in Table \ref{tab:comp_acc}. All benchmarking methods are evaluated in two metrics \cite{qi2017pointnet}: the average of per-class accuracy (class-avg) and the overall accuracy of all scans. Among DL-based methods, we see 4.5\% overall accuracy improvement from PointNet in 2017 to Point Transformer in 2021 \cite{zhao2021point}. Nevertheless, the improvement comes at the price of a substantially larger model size and a significantly higher complexity. Among GL-based methods, Green-PointHop achieves 86\% class-avg and 90.6\% overall accuracy, which are second (but close) to PointHop++. Since Green-PointHop is a single-scale method with only one hop, we report the classification accuracy of PointHop and PointHop++ using only one hop in the same table for comparison. Clearly, both of them have a large performance gap with respect to Green-PointHop.  It demonstrates the effectiveness of aggregating complementary geometric information through multiple oriented cones. By comparing DL- and GL-based methods, the performance of Green-PointHop is comparable with those of PointNet and PointNet++. 

%%%%%%%%%%%%%%%%%%%%%%%%%%%%%%%%%%%%%%%%%%%%%%%%%%%%%%%%%%%%%%%%%%%%%%
\begin{table}[htbp!]
\centering
\caption{Classification performance comparison of representative DL- and GL-based point cloud object classification methods on ModelNet40.}
\label{tab:comp_acc}
\renewcommand\arraystretch{1.1}
% \resizebox{\columnwidth}{!}{
%\resizebox*{\textwidth}{!}{
\begin{tabular}{c|c|c|c|c} \hline
\multirow{2}*{Learning Scheme} & \multicolumn{2}{c|}{\multirow{2}*{Method}} & \multicolumn{2}{c}{Accuracy (\%)} \\ \cline{4-5}
& \multicolumn{2}{c|}{} & class-avg & overall  \\ \hline
\multirow{8}*{Deep Learning} & \multicolumn{2}{c|}{PointNet \cite{qi2017pointnet}} & 86.2 & 89.2 \\
& \multicolumn{2}{c|}{PointNet++ \cite{qi2017pointnet++}} & - & 90.7 \\
& \multicolumn{2}{c|}{PointCNN \cite{li2018pointcnn}} & 88.1 & 92.5 \\
& \multicolumn{2}{c|}{PointConv \cite{wu2019pointconv}} & - & 92.5 \\
& \multicolumn{2}{c|}{DGCNN \cite{wang2018dynamic}} & 90.2 & 92.9 \\ 
& \multicolumn{2}{c|}{KPConv \cite{thomas2019kpconv}} & - & 92.9 \\ 
& \multicolumn{2}{c|}{PCT \cite{guo2021pct}} & - & 93.2 \\ 
& \multicolumn{2}{c|}{Point Transformer \cite{zhao2021point}} & 90.6 & 93.7 \\ \hline
\multirow{6}*{Green Learning} 
& \multirow{2}*{4 hops} & PointHop \cite{zhang2020pointhop} & 84.4 & 89.1 \\
& & PointHop++ \cite{zhang2020pointhop++} & 87 & 91.1 \\ \cline{2-5}
& \multirow{4}*{1 hop} & PointHop & 44.8 & 60.7 \\
& & PointHop++ & 49.7 & 64.5 \\ 
& & Green-PointHop (Ours) & 86 & 90.6 \\\hline
\end{tabular}%}
\end{table}
%%%%%%%%%%%%%%%%%%%%%%%%%%%%%%%%%%%%%%%%%%%%%%%%%%%%%%%%%%%%%%%%%%%%%%

\subsubsection{Model and Time Complexities}\label{model_complex}

%%%%%%%%%%%%%%%%%%%%%%%%%%%%%%%%%%%%%%%%%%%%%%%%%%%%%%%%%%%%%%%%%%%%%%
\begin{table}[htbp!]
\centering
\caption{Complexity comparison of representative DL- and GL-based methods in four measures: 1) hardware configuation, 2) FLOPs in inference, 3) training and inference time, and 4) the number of model parameters (i.e., model size), where `M' stands for the unit of a million.} \label{tab:comp_complexity}
\renewcommand\arraystretch{1.1}
\resizebox{\columnwidth}{!}{
% \resizebox*{\textwidth}{!}{
\begin{tabular}{c|c|c|cc|ccc} \hline
\multirow{2}*{Method} & Hardware & \multirow{2}*{FLOPs (M)} & 
\multicolumn{2}{c|}{Time (hr, ms)} & \multicolumn{3}{c}{Parameter No. (M)} \\ \cline{4-8}
& Configuration & & Training & Inference & Filter & Classifier & Total \\ \hline
PointNet \cite{qi2017pointnet} & \multirow{3}*{GPU} & 957 (416X) & 7 & 10 & - & - & 3.48 (56.13X) \\ 
PointNet++ \cite{qi2017pointnet++} & & 3136 (1363X) & 7 & 14 & - & - & 1.48 (23.87X) \\ 
DGCNN \cite{wang2018dynamic} & & 2768 (1203X) & 21 & 154 & - & - & 2.63 (42.42X)\\ \hline
PointHop \cite{zhang2020pointhop} & \multirow{3}*{CPU} & 7.7 (3X) & 0.33 & 108 & 0.037 & - & - \\ 
PointHop++ \cite{zhang2020pointhop++} & & 4.0 (2X) & 0.42 & 97 & 0.009 & 0.15 & 0.159 (2.56X)\\
Green-PointHop (Ours) & & 2.3 (1X) & 0.08 & 23 & 0.002 & 0.06 & 0.062 (1X) \\ \hline
\end{tabular}%
}
\end{table}
%%%%%%%%%%%%%%%%%%%%%%%%%%%%%%%%%%%%%%%%%%%%%%%%%%%%%%%%%%%%%%%%%%%%%%  

We compare the model and time complexities of representative DL-based and GL-based methods in Table \ref{tab:comp_complexity}. We adopt different hardware platforms for DL- and GL-based methods due to their different algorithmic nature. As reported in the second column, DL-based methods are trained on a single GeForce GTX TITAN X GPU while GL-based methods are trained on a 24-thread Intel(R) Xeon(R) CPU. It takes several hours to train DL models. In contrast, it takes around 20 minutes and 25 minutes to train PointHop and PointHop++, respectively. The new Green-PointHop model completes its training within 5 minutes. Although we conduct Green-PointHop's inference on CPU, its inference time (23 ms) is close to that of PointNet and PointNet++ using GPU (i.e., 10 ms and 14 ms, respectively). 

One platform-independent computational complexity measure is the number of floating-point operations (FLOPs). The number of FLOPs is proportional to power consumption as well as carbon footprint. In the 3rd column, we report the numbers of FLOPs in the unit of millions and show their relative ratios with respect to that of Green-PointHop inside the parentheses. Green-PointHop has around 2.3 million FLOPs, which is about one half and one third of PointHop and PointHop++, respectively. For the FLOP number of DL-based methods, we cite the data in \cite{chen2020go}. As shown in Table \ref{tab:comp_complexity}, the FLOP numbers of DL-based methods are about three orders of magnitude of that of Green-PointHop, which shows power efficiency of Green-PointHop.

Finally, for model size comparison, we list the numbers of model parameters in the last column of the table. For GL-based methods, besides the total number of model parameters, we report the parameter numbers in two modules separately: 1) the filters used in unsupervised feature learning, and 2) the LLSR classifier. Green-PointHop has significantly fewer parameters in these two modules than PointHop and PointHop++ because of its single-hop architecture. The total number of parameters of Green-PointHop is only 62K, which can easily fit into a 256k-byte cache if each parameter is stored in four bytes. 

\subsubsection{Ablation Study} \label{ablation_study}

We conduct extensive experiments with various settings of Green-PointHop on ModelNet40 to shed light on several design choices. 

%%%%%%%%%%%%%%%%%%%%%%%%%%%%%%%%%%%%%%%%%%%%%%%%%%%%%%%
\begin{figure}[htbp!]
\centerline{\includegraphics[width=3.5in]{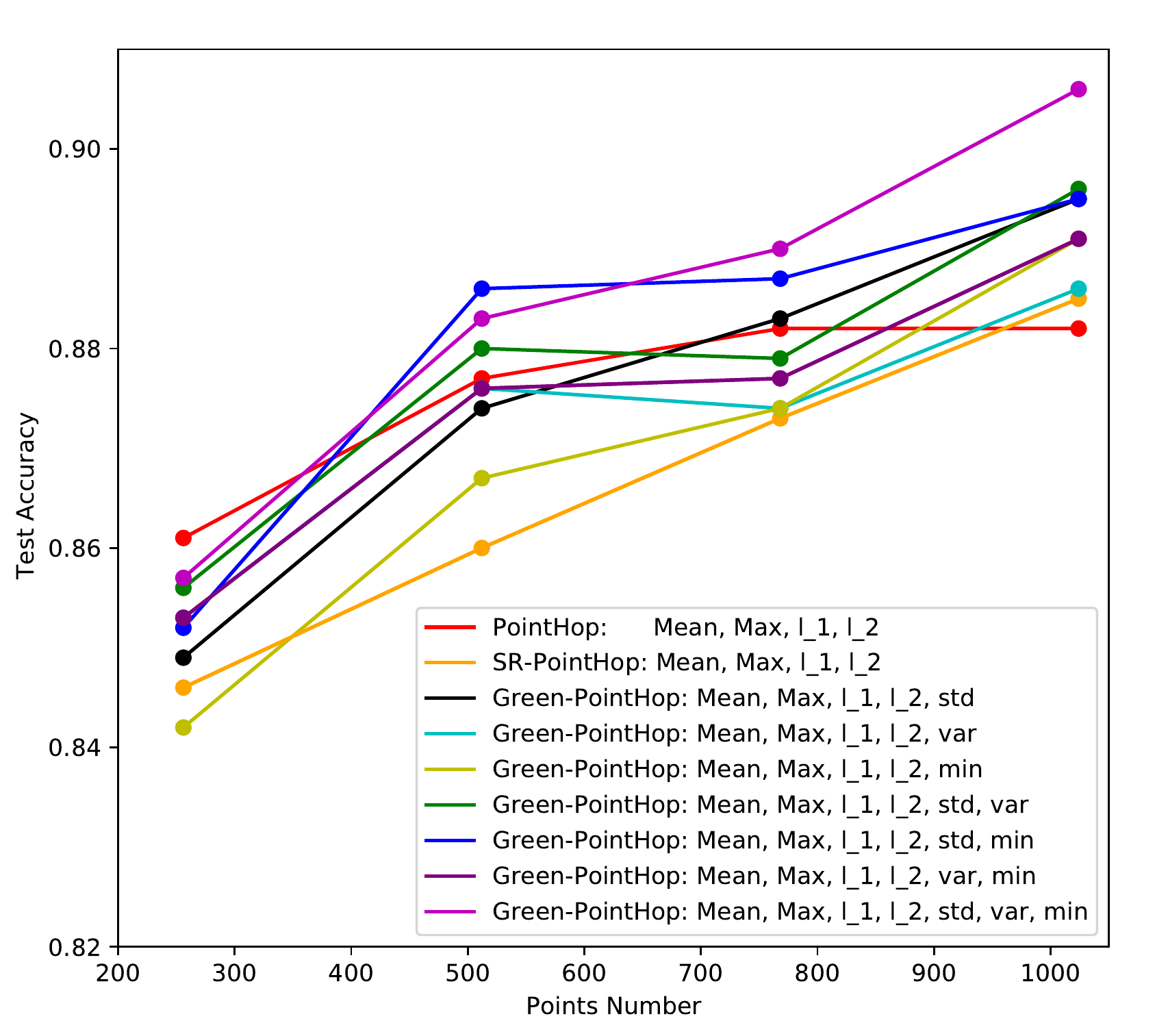}}
\caption{The plot of Green-PointHop's overall classification accuracy as a function of the aggregating point number under eight different settings those adopt a subset of seven aggregation quantities (i.e., max, min, mean, $L_1$, $L_2$, std and variance values) of selected points.}\label{fig:aggregation}
\end{figure}
%%%%%%%%%%%%%%%%%%%%%%%%%%%%%%%%%%%%%%%%%%%%%%%%%%%%%%%%

{\bf Global Aggregation Schemes.} We first study the effect of using different global aggregation schemes on the overall classification accuracy in Fig. \ref{fig:aggregation}. The x-axis of the figure indicates the aggregation across four sets of points (i.e., 256, 512, 768, and 1024 points). The first three sets are randomly sampled from a total of 1024 points. The general trend is that the test accuracy improves as the number of sampled points increases, being attributed to a better coverage of underlying point cloud scans. 

As to the number of aggregation quantities, we begin with the best setting of PointHop, which is an ensemble of max, mean, $L_1$ and $L_2$ four quantities. Under this setting, Green-PointHop has worse performance than PointHop except for 1024 points. To improve the performance, we calculate three more quantities as the new aggregation schemes. They are the min, the standard deviation and the variance across a selected set of spatial points. They offer new ways to summarize the global information obtained at the first stage in Fig. \ref{fig:pipeline}. We compare the test accuracy of eight aggregation cases of Green-PointHop in Fig. \ref{fig:aggregation}. As shown in the figure, the ensemble of all seven aggregation schemes gives the best performance. The robustness of Green-PointHop is demonstrated by good test performance with fewer points in point cloud object scans. Although Green-PointHop has one scale only, it offers the best performance for cases of 768 and 1024 points and the second best performance for cases of 256 and 512 points. 

{\bf Sizes of Nearest Neighborhood.} Next, we investigate the effect of the number of nearest neighbors (or the number of $K$ in the KNN search) adopted in the first stage of Fig.  \ref{fig:pipeline}. We examine $K=16$, 32 and 64 three cases in the first three columns of Table \ref{tab:ablation_study}. By focusing on the last three rows, we see $K=32$ gives the best result when other conditions are equal. It outperforms the cases of $K=16$ and $K=64$ by 1.5\% and 0.85\%, respectively. Thus, we successfully reduce $K$ from 64 in PointHop to 32 in Green-PointHop. This helps reduce the memory size and time complexity. 

{\bf Effects of Global and Local Cone and Inverted Cone Aggregations.} When $K=32$, we see from Table \ref{tab:ablation_study} that Green-PointHop has the highest accuracy (90.64\%) by concatenating features obtained from the global aggregation, the local cone aggregation, and the local inverted cone aggregation. It validates that the proposed cone and inverted cone aggregations can capture more shape information of objects than the simple global aggregation. This study shows the power of geometric aggregations of local representations. 

%%%%%%%%%%%%%%%%%%%%%%%%%%%%%%%%%%%%%%%%%%%%%%%%%%%%%%%%%%%%%%%%%%
\begin{table}[htbp!]
\centering
\caption{Ablation Study on the number of nearest neighbors and different combinations of global and local aggregations. For the number of nearest neighbors, only the results of three commonly used $K$ are given for comparison.  For the column of aggregation, cones and inverted cones are tested separately. } \label{tab:ablation_study}
\renewcommand\arraystretch{1.1}
\newcommand{\tabincell}[2]{\begin{tabular}{@{}#1@{}}#2\end{tabular}}
% \resizebox{\columnwidth}{!}{
\begin{tabular}{ccc|ccc|c} \hline 
\multicolumn{3}{c|}{Nearest Neighbor \# (K)} & \multicolumn{3}{c|}{Aggregation} & \multirow{2}{*}{\tabincell{c}{Overall \\ Accuracy (\%)}}  \\ \cline{1-6}
\hspace*{1.4mm} 16 \hspace*{1.4mm} & \hspace*{1.4mm} 32 \hspace*{1.4mm} & \hspace*{1.4mm} 64 \hspace*{1.4mm} & Global & Cone & Inverted Cone & \\ \hline 
& \checkmark & & \checkmark & & & 72.37 \\ 
& \checkmark & & & \checkmark & & 83.75 \\ 
& \checkmark & & & & \checkmark & 87.1 \\ 
& \checkmark & & \checkmark & \checkmark & & 85.94 \\ 
& \checkmark & & \checkmark & & \checkmark & 88.49 \\ 
& \checkmark & & & \checkmark & \checkmark & 89.75 \\ 
& \checkmark & & \checkmark & \checkmark & \checkmark & \bf{90.64} \\  
\checkmark & & & \checkmark & \checkmark & \checkmark & 89.14 \\ 
& & \checkmark & \checkmark & \checkmark & \checkmark & 89.79 \\ \hline 
\end{tabular}%}
\end{table}
%%%%%%%%%%%%%%%%%%%%%%%%%%%%%%%%%%%%%%%%%%%%%%%%%%%%%%%%%%%%%%%%%%

\subsection{ScanObjectNN Dataset}\label{ScanObjectNN}

We conduct experiments on a real world dataset, ScanObjectNN with object only, to show the generizability and effectiveness of Green-PointHop in this subsection. For fair comparison, we perform the same data augmentation, including random rotation and per-point jitter, as done in \cite{uy2019revisiting}. We compare the classification accuracy results of several DL-based methods and Green-PointHop in Table \ref{tab:comp_acc_scanobj}. Green-PointHop achieves 77.5\% per-class accuracy and 79.3\% overall accuracy. These numbers are comparable with PointNet and SpiderCNN. Green-PointHop has a performance gap against more advanced DL-based methods, ranging from 5\% to 6.9\% in the overall classification accuracy. Since the discussion on model/time complexities in Sec. \ref{model_complex} still holds here, Green-PointHop offers a different tradeoff between complexities and effectiveness. 

%%%%%%%%%%%%%%%%%%%%%%%%%%%%%%%%%%%%%%%%%%%%%%%%%%%%%%%%%%%%%%%%%%%%%%
\begin{table}[htbp!]
\centering
\caption{Comparison of classification results on ScanObjectNN.}
\renewcommand\arraystretch{1.1}
% \resizebox{\columnwidth}{!}{
\begin{tabular}{c|c c} \hline
\multirow{2}*{Method} & \multicolumn{2}{c}{Accuracy (\%)} \\ \cline{2-3}
& class-avg & overall  \\ \hline
{3DmFV \cite{ben20183dmfv}} & 68.9 & 73.8 \\
{PointNet \cite{qi2017pointnet}} & 74.4 & 79.2 \\
{SpiderCNN \cite{xu2018spidercnn}} & 77.4 & 79.5\\
{PointNet++ \cite{qi2017pointnet++}} & 82.1 & 84.3 \\
{PointCNN \cite{li2018pointcnn}} & 83.3 & 85.5 \\
{DGCNN \cite{wang2018dynamic}} & 84.0 & 86.2 \\ 
{Green-PointHop} (Ours) & 77.5 & 79.3 \\\hline
\end{tabular}%}
\label{tab:comp_acc_scanobj}
\end{table}
%%%%%%%%%%%%%%%%%%%%%%%%%%%%%%%%%%%%%%%%%%%%%%%%%%%%%%%%%%%%%%%%%%%%%%

We have an interesting observation by comparing the performance of PointNet and Green-PointHop. The per-class accuracy of Green-PointHop is 3.1\% higher than that of PointNet while their overall accuracy results are about the same. It means that Green-PointHop has more balanced performance across different classes. To check this, we compare the per-class accuracy of Green-PointHop and PointNet in Table \ref{tab:SR_pointnet}. Green-PointHop achieves higher accuracy in 11 classes and lower accuracy in the remaining 4 classes (cabinet, desk, door, and table).  

%%%%%%%%%%%%%%%%%%%%%%%%%%%%%%%%%%%%%%%%%%%%%%%%%%%%%%%%%%%%%%%%%%%%%%
\begin{table*}[!htbp]
\centering
\caption{Comparison of per-class classification accuracy of PointNet and Green-PointHop on the ScanObjectNN dataset.}\label{tab:SR_pointnet}
\renewcommand\arraystretch{1.1}
\begin{tabular}{c|cccccccccc} \hline
Method & Bag & Bin & Box & Cabinet & Chair \\ \hline
PointNet & 47.1 & 80.0 & 35.7 & \bf{80.0} & 93.6 \\
Green-PointHop (Ours) & \bf{70.6} & \bf{90.0} & \bf{46.4} & 65.3 & \bf{96.2} \\ \hline
& Desk & Display & Door & Shelf & Table  \\ \hline
PointNet & \bf{86.7} & 73.8 & \bf{97.6} & 81.6 & \bf{88.9}  \\
Green-PointHop (Ours) & 56.7 & \bf{83.3} & 92.9 & \bf{83.7} & 79.6  \\ \hline
& Bed & Pillow & Sink & Sofa & Toilet \\ \hline
PointNet & 59.1 & 76.2 & 54.2 & 85.7 & 76.5 \\
Green-PointHop (Ours) & \bf{81.8} & \bf{81.0} & \bf{58.3} & \bf{88.1} & \bf{88.2} \\ \hline        
\end{tabular}
\end{table*}
%%%%%%%%%%%%%%%%%%%%%%%%%%%%%%%%%%%%%%%%%%%%%%%%%%%%%%%%%%%%%%%%%%%%%%

\begin{figure}[htbp!]
\centerline{\includegraphics[width=0.78\textwidth]{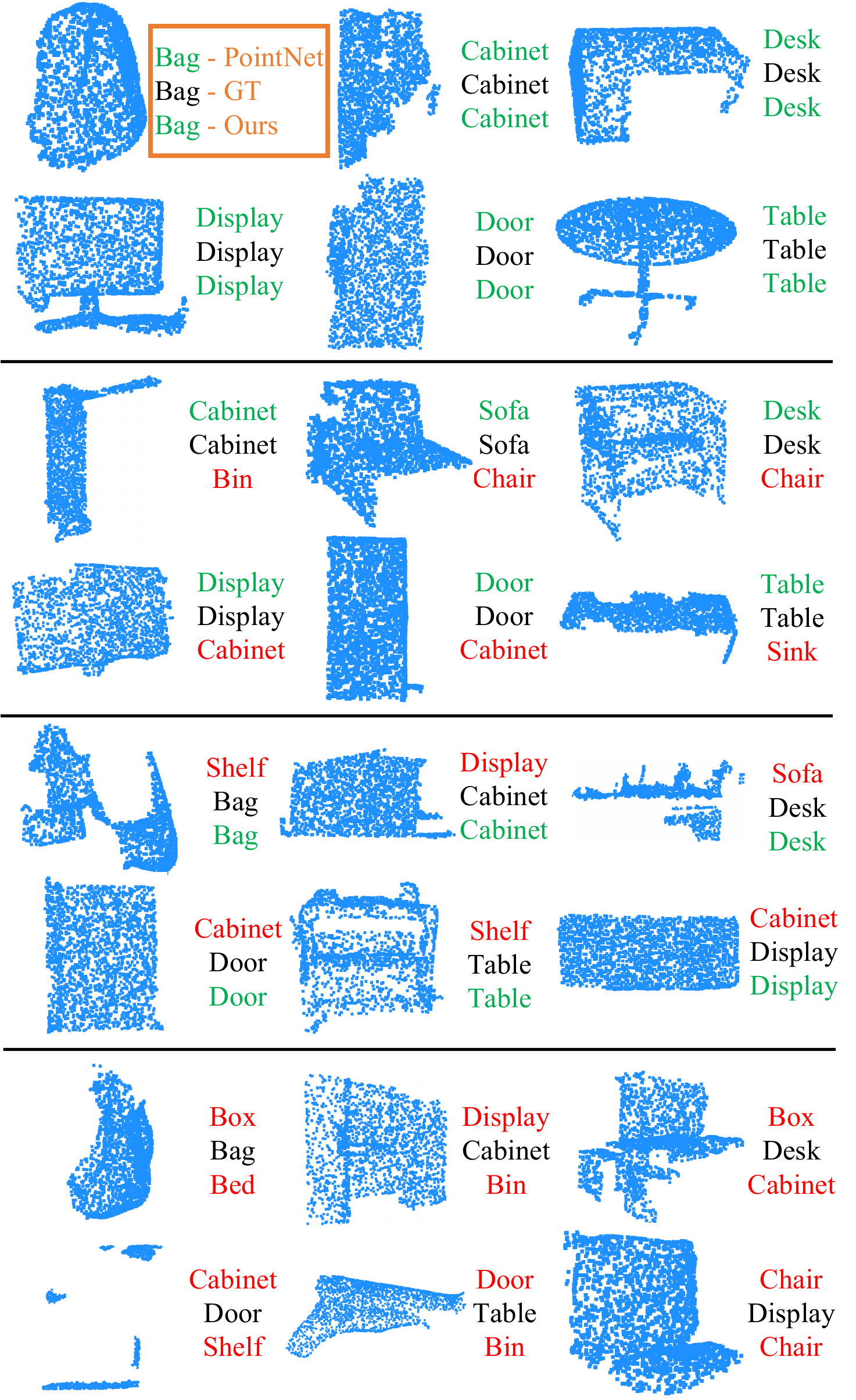}}
\caption{Prediction error analysis for ScanObjectNN. Each object is visualized with its GT label in the middle while the labels predicted by PointNet and Green-PointHop are, respectively, shown above and below for comparison. Also, wrong and correct predictions are shown in red and green, respectively.} \label{fig:err_analysis}
\end{figure}

To understand it further, we show some objects in the ScanObjectNN dataset with ground truth (GT) labels and predictions made by PointNet and Green-PointHop in Fig. \ref{fig:err_analysis}, where labels and predictions are annotated for error analysis. We show six examples for each of the following four cases: 1) both predict correctly (the top group); 2) only PointNet predicts correctly (the second group); 3) only Green-PointHop predicts correctly (the third group); 4) neither predicts correctly (the bottom group). 

Visualization of these objects helps us understand the challenges of the real world dataset. One main obstacle is that the shape of many objects is incomplete (with partial surfaces only). These incomplete objects are difficult to classify as shown in Fig. \ref{fig:err_analysis}. For example, the door located in the bottom left corner of the figure has only some borders. Even humans have difficulty in recognizing it. Also, we observe that door, cabinet and display form a confusing group. They share a similar geometric shape -- a flat surface. Only some details on surfaces or edges can make them distinctive. However, these discriminant parts could be missing. 

%%%%%%%%%%%%%%%%%%%%%%%%%%%%%%%%%%%%%%%%%%%%%%%%%%%%%%%
%%%%%%%%%%%%%%%%%%%%%%%%%%%%%%%%%%%%%%%%%%%%%%%%%%%%%%%%

\section{Conclusion and Future Work} \label{sec:conclusion}

An extremely lightweight machine learning model, called Green-PointHop, was proposed for point cloud object classification in this work. Green-PointHop simplifies the PointHop model by reducing the hop number from four to one. To enrich the representations of a point cloud object, Green-PointHop exploited aggregated features obtained from points in the whole object and its partial views. This new idea has a clear geometric interpretation. Furthermore, its model size and computation complexity can be significantly reduced since it has filtering operations at a single scale. Experiments on ModelNet40 and ScanObjectNN datasets showed that the classification performance of Green-PointHop is comparable with PointNet and PointNet++. 

There are several research topics worth future exploration. First, it is interesting to extend Green-PointHop to solve the point cloud segmentation and registration problems. Second, it is important to have a better understanding on the pooling operation among spatial points and aggregations of filter responses across points in spatial regions. The understanding will help guide the selection of optimal hyper-parameters of Green-PointHop. 

\section*{Acknowledgment}

This work was supported by a research grant from Tencent. The authors acknowledge the Center for Advanced Research Computing (CARC) at the University of Southern California for providing computing resources that have contributed to the research results reported in this publication. 

\bibliographystyle{unsrt}  
\bibliography{ref}

\end{document}